\documentclass[11pt,a4paper]{article}
\usepackage{times,latexsym}
\usepackage{url}
\usepackage[T1]{fontenc}

\usepackage[acceptedWithA]{tacl2021v1}
\usepackage{tacl2021v1}

\usepackage{xspace,mfirstuc,tabulary}

\usepackage{graphicx}
\usepackage{inconsolata}
\usepackage{array}
\usepackage{mathrsfs}
\usepackage{amsthm}
\usepackage{amsmath}
\usepackage{amsfonts}
\usepackage{algorithm}
\usepackage{algorithmicx}
\usepackage[noend]{algpseudocode}
\usepackage{multirow}
\usepackage{booktabs}
\usepackage{color}
\usepackage{arydshln}

\newtheorem{definition}{Definition}

\newif\iftaclinstructions
\taclinstructionsfalse
\iftaclinstructions
\renewcommand{\confidential}{}
\renewcommand{\anonsubtext}{(No author info supplied here, for consistency with
TACL-submission anonymization requirements)}
\newcommand{\instr}
\fi

\iftaclpubformat

\else

\fi

\title{Consistent Multi-Granular Rationale Extraction for Explainable Multi-hop Fact Verification}

\author{Jiasheng Si$^1$ \ \ \ \  Yingjie Zhu$^1$ \ \ \ \  Deyu Zhou\Thanks{corresponding author} \ \ \ \   \\
School of Computer Science and Engineering, Key Laboratory of Computer Network\\
	and Information Integration, Ministry of Education, Southeast University, China \\
\texttt{\{jasenchn, yj\_zhu, d.zhou\}@seu.edu.cn}}

\date{}

\begin{document}
\maketitle
\begin{abstract}
  The success of deep learning models on multi-hop fact verification has prompted researchers to understand the behavior behind their veracity. One possible way is erasure search: obtaining the rationale by entirely removing a subset of input without compromising the veracity prediction. Although extensively explored, existing approaches fall within the scope of the single-granular (tokens or sentences) explanation, which inevitably leads to explanation redundancy and inconsistency. To address such issues, this paper explores the viability of multi-granular rationale extraction with consistency and faithfulness for explainable multi-hop fact verification. In particular, given a pretrained veracity prediction model, both the token-level explainer and sentence-level explainer are trained simultaneously to obtain multi-granular rationales via differentiable masking. Meanwhile, three diagnostic properties (fidelity, consistency, salience) are introduced and applied to the training process, to ensure that the extracted rationales satisfy faithfulness and consistency. Experimental results on three multi-hop fact verification datasets show that the proposed approach outperforms some state-of-the-art baselines.
\end{abstract}

\section{Introduction}
\label{sec:introduction}

Computational fact checking approaches typically explore neural models to verify the truthfulness of a claim by reasoning over multiple pieces of evidence~\citep{hover,politihop}. 
However, few methods have been devoted to acquiring explanations for these systems, which weakens user trust in the prediction and prohibits the discovery of artifacts in datasets~\citep{survey2020, survey2022,janizek2021}. 
In this work, 
we explore \textbf{post hoc interpretability}, aiming to explain the veracity prediction of a multi-hop fact verification model and reveal how the model arrives at the decision by retaining subsets of the input (i.e., \textit{rationale}).

To understand the behavior of a model with a certain prediction, a classical way to perform explaining is  \textbf{erasure search}~\citep{DBLP:journals/corr/LiMJ16a, feng-etal-2018-pathologies, decao2020, atanasov2022, si2022exploring}, 
an approach wherein rationale is obtained by searching for a maximum subset of the input (e.g., tokens, sentences) that can be completely removed from the input without affecting the veracity prediction\footnote{Following~\citet{si2022exploring}, we define \textit{true evidence} as \textit{rationale} rather than \textit{noise evidence}.}.
This removal perturbation to the input guarantees the decorrelation of discarded features with veracity prediction of the model, 
in contrast to the intrinsic approaches (e.g., attention-based methods) that cannot ensure the ignoring of low-scoring input features~\citep{generating-fact2020, pubhealth-2020, predictagain2021, claimdissector2022}. 

\begin{figure}[pt]
    \centering
    \includegraphics[width=7.7cm]{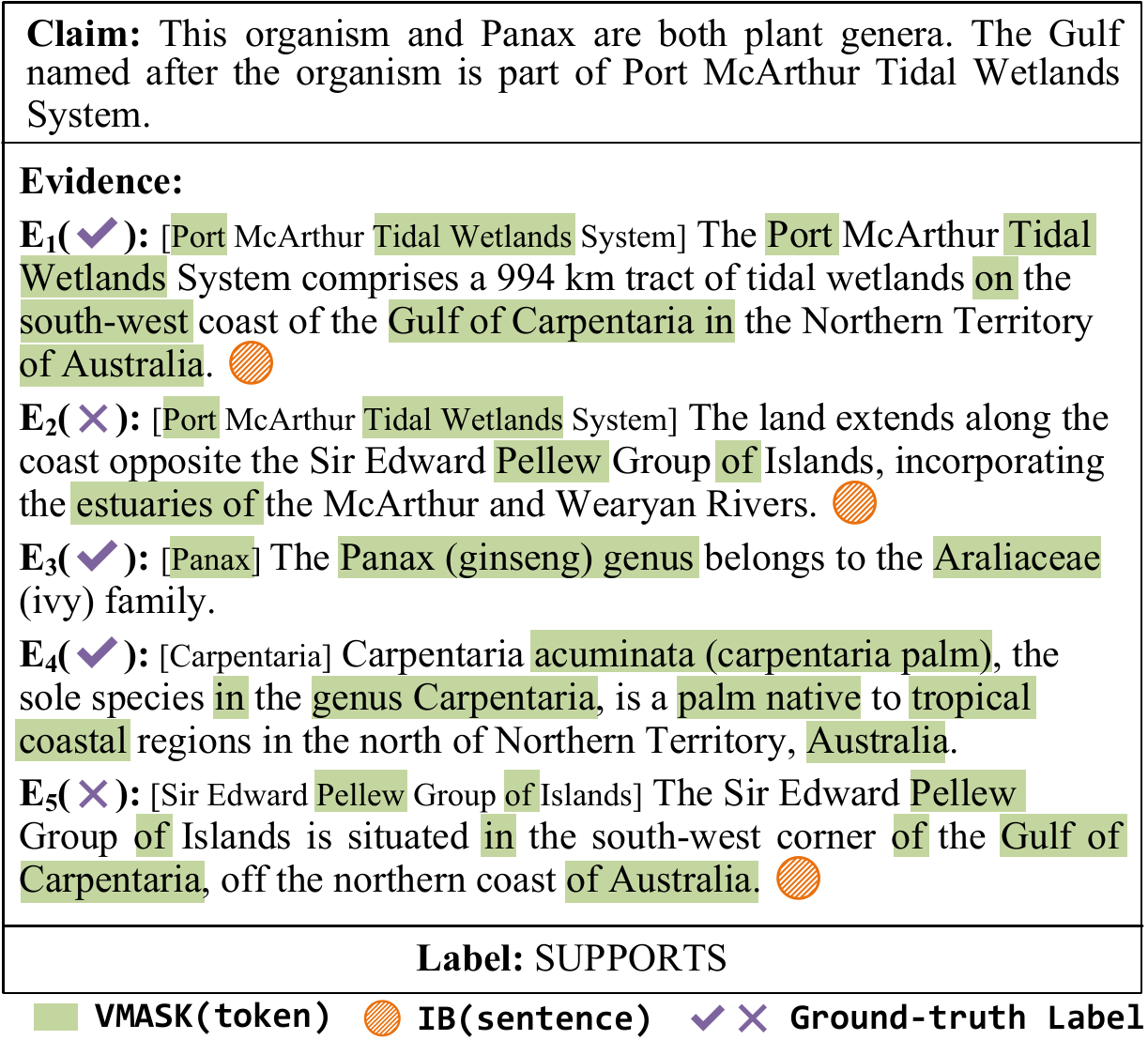}
    \caption{
    An example in the HoVer dataset marked with sentence-level rationales extracted by Information Bottleneck (IB)~\citep{informationbottleneck} and token-level rationales extracted by VMASK~\citep{vmask2020}.    
    }
    \label{fig:introduction}
\end{figure}

Existing explanation approaches for multi-hop fact verification based on erasure searching can be categorized into sentence-level rationale extraction based~\citep{informationbottleneck,atanasov2022, si2022exploring} and token-level rationale extraction based~\citep{lime, shap_nips, intgrad, decao2020, vmask2020, evarm2022}.
Despite extensive exploration, 
we found that the rationales extracted at the sentence level are too coarse and might contain irrelevant and redundant tokens to the claim 
(e.g., ``\textit{994km tract of tidal wetlands}'' in \textit{E1} and ``\textit{Gulf of Carpentaria}'' in \textit{E5} in Figure~\ref{fig:introduction}.),
which impairs the ability to shield the effect of noise evidence. 
Obviously, this issue can be overcome by extracting rationale at the token level.
However, current token-level rationale extraction methods lack the capability to discern the true evidence and noise evidence. This poses a major challenge to explain the multi-hop fact verification model. 
Extensive redundant and confusing tokens will inevitably be extracted as rationales from the noise evidence,
thus inducing \textit{inconsistency} between the extracted tokens with the true evidence
(e.g., the token rationales should not be extracted from \{\textit{E2}, \textit{E5}\} in Figure~\ref{fig:introduction}).
This results in under-aggressive pruning that is unable to reflect the intrinsic information the model relies on to arrive at the decision.
Therefore, this paper seeks to explore a feasible way to extract the \textit{``right tokens''} from the \textit{``right sentences''} (i.e., we aim to only retain the task-relevant tokens contained in \{\textit{E1}, \textit{E3}, \textit{E4}\}.).
For this purpose, we propose a novel paradigm to yield indicative token rationales by extracting multi-granular rationales with regularization.

Although promising,
follow-up questions then arise:
(i) \textit{How to extract the multi-granular rationales simultaneously for the veracity prediction?}
(ii) \textit{How to ensure the faithfulness~\citep{fresh} and consistency of the multi-granular rationales?}
In this paper, we give affirmative answers to the questions and offer a novel \textbf{C}onsistent m\textbf{U}lti-granular \textbf{R}ationale \textbf{E}xtraction (CURE) approach for explainable multi-hop fact verification.
The \textbf{core idea} of our CURE is that both the token-level explainer and the sentence-level explainer are learned simultaneously, with the desire to extract the \textit{consistent} multi-granular rationales and make them \textit{faithful} toward the verification.
It ensures the mutual effect between the information of retained tokens and sentences and produces the indicative token rationales.
In specific,
given a pretrained multi-hop fact verification model,
we first train two parameterized explainers to generate mask vectors for each token and sentence to indicate which token or sentence is necessary or can be discarded,
based on the intermediate hidden representation of the Transformer-XH ~\citep{transformer-xh}. 
Then, the two learnable mask vectors are intersected and induced back into the input to remove the irrelevant tokens and sentences.
Meanwhile, a capsule network~\citep{capsule} is used to aggregate the retained features by intervening on coupling coefficients with the sentence mask. In addition, three diagnostic properties are introduced as guidance to regularize rationale extraction, (i) Fidelity to constrain the \textit{faithfulness} of rationales;
(ii) Consistency to increase the \textit{consistency} between the multi-granular rationales; 
(iii) Salience to guide the rationale extraction with predefined salience score.

In a nutshell, our main contributions can be summarized as follows: 
(I) We for the first time explore the multi-granular rationale extraction for the explainable multi-hop fact verification. 
(II) Three diagnostic properties are designed and applied to regularize rationale extraction to achieve faithfulness and consistency. 
(III) Experiments on three multi-hop fact verification datasets are conducted to validate the superiority of our approach.

\begin{figure*}[t]
    \centering
    \includegraphics[width=13cm]{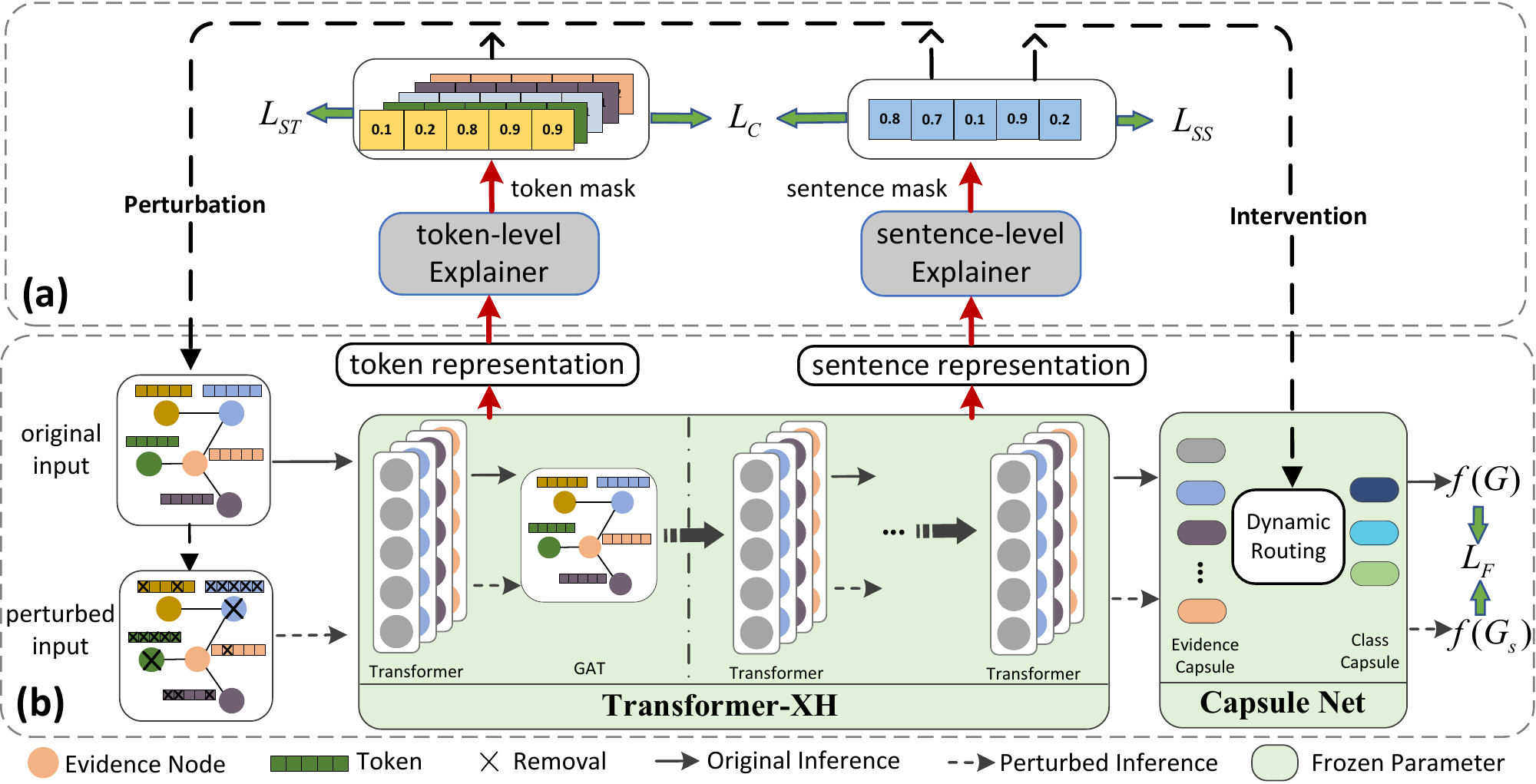}
    \caption{The overall architecture of our CURE. (a): the multi-granular rationale extraction, (b): the veracity prediction model}
    \label{fig:method}
\end{figure*}

\section{Preliminaries}
\paragraph{Task}
Following \citet{liu-etal-2020-fine}, 
given a claim $\boldsymbol{c}$ with associated evidence $\{\boldsymbol{e_1},\boldsymbol{e_2},\dots,\boldsymbol{e_n}\}$, 
we construct a fully connected input graph $G=(X,A)$, 
where $n$ is the number of evidence,
$\boldsymbol{x_i}\in X$ denotes the evidence node by concatenating the evidence text $\boldsymbol{e_i}$ with the claim $\boldsymbol{c}$. 
We aim to jointly extract multi-granular rationales with the desire of faithfulness and consistency for explainable multi-hop fact verification, 
i.e., sentence-level rationales $R_s = (X_s, A_s\in\mathbb{R}^{n\times n})$ and token-level rationales $\boldsymbol{r}=\{\boldsymbol{r}_i\subset\boldsymbol{e}_i\}|_{i=0}^n$, where $\boldsymbol{r}_i=\{\boldsymbol{t}_{i,j}|\boldsymbol{t}_{i,j} \in \boldsymbol{e}_i\}$.
We only extract token rationales from $\boldsymbol{e}_i$ and denote $|\boldsymbol{x}_i|$ as the number of tokens in $i$th evidence node.

\begin{definition}
    \label{def:faithful}
    (\textit{Faithfulness}) $R_s$ and $\boldsymbol{r}$ are multi-granular faithful to their corresponding prediction $Y$ if and only if $Y$ rely entirely on $G_R=(\{\boldsymbol{x}_i\cap \boldsymbol{r}_i\mid \boldsymbol{x}_i \in X_s\}, A_s)$.
\end{definition}

\begin{definition}
    \label{def:non-redundant}
    (\textit{Consistency}) $R_s$ and $\boldsymbol{r}$ are multi-granular consistent to their corresponding prediction $Y$ if and only if satisfying
    \begin{equation}
    \label{eq:non-redundancy}
        \sum_{\boldsymbol{x_i} \in X_s} |\boldsymbol{x_i}\cap \boldsymbol{r}_i| \le \epsilon, 
        \sum_{\boldsymbol{x'_i} \in X\setminus X_s} |\boldsymbol{x'_i}\cap \boldsymbol{r}_i| \rightarrow 0 ,
    \end{equation}
    where $\epsilon$ is the maximum expected sparsity of token-level rationales.
    $X\setminus X_s$ denotes the complementary subset of $X_s$.
\end{definition}

\section{Method}
\label{sec.method}

We now describe the proposed methods in detail, which includes the architectures:
(i) a veracity prediction model (shown in Figure~\ref{fig:method}(b)),
(ii) multi-granular rationale extraction (shown in Figure~\ref{fig:method}(a)),
and the terms we optimize:
(iii) the diagnostic properties,
(iv) the optimization.

\subsection{Veracity Prediction}
\label{subsec.veracity_prediction}
For the multi-hop fact verification model $f(\cdot)$, as shown in Figure~\ref{fig:method}(b),
we employ the classical veracity model Transformer-XH combined with capsule network as illustrated in~\citet{si2021}.

\textbf{Semantic encoder} 
Given a graph $G=(X, A)$, 
a Transformer layer is first applied to the node $X$ to obtain the token representation $\boldsymbol{h}=\langle \boldsymbol{h}_0,\boldsymbol{h}_1,...,\boldsymbol{h}_n\rangle$ for each evidence, 
where $\boldsymbol{h}_i=\langle \boldsymbol{h}_{i,0}, \boldsymbol{h}_{i,1},...,\boldsymbol{h}_{i,|\boldsymbol{x}_i|} \rangle$,
$\boldsymbol{h}_{i,j}$ denotes the $j$th token representation in $i$th evidence.
Then, a GAT layer is applied to the \textit{[CLS]} token representation to propagate the message exchange among all evidence along the edges, 
i.e., $\tilde{\boldsymbol{h}}_{i,0}|_{i=1}^n = GAT(\boldsymbol{h}_{i,0}|_{i=1}^n)$.
The updated representation thus is obtained, $\boldsymbol{h}_i=\langle \tilde{\boldsymbol{h}}_{i,0}, \boldsymbol{h}_{i,1},...,\boldsymbol{h}_{i,|\boldsymbol{x}_i|}\rangle$,
where $\tilde{\boldsymbol{h}}_{i,0}$ denotes the sentence representation for $i$th evidence.
By stacking $L$-layers of Transformer with GAT, we get the representation $\boldsymbol{H}=\langle \boldsymbol{h}^0,\boldsymbol{h}^1,...,\boldsymbol{h}^L\rangle$, where $\boldsymbol{h}^0=X$.

\textbf{Aggregator}
We use the capsule network to aggregate the information among all the evidence by taking sentence representation $\tilde{\boldsymbol{h}}_{i,0}^L|_{i=0}^n$ as the evidence capsule and label as the class capsule.
It permits us to further eliminate the effect of non-rationale for veracity prediction.
The capsule loss is used to optimize the veracity prediction model.

\subsection{Multi-granular Rationale Extraction}
\label{subsec.explainer}
Our CURE relies on \textit{erasure search},
retaining minimal but sufficient multi-granular rationales while maintaining the original veracity~\citep{decao2020}.
As shown in Figure~\ref{fig:method}(a),
we propose two parameterized explainers to both generate the binary mask vectors at the token level and sentence level, indicating the absence or presence of each token or sentence.

Taking the hidden representation $\boldsymbol{H}$ from multiple layers in Transformer-XH as input,
for the token-level explainer,
we employ a shallow interpreter network $g_t(\cdot)$ (i.e., one-hidden-layer MLP network) to yield binary token mask vectors $\boldsymbol{z}=\{\boldsymbol{z}_i\}|_{i=0}^n$ conditioned on the token representation,
where $\boldsymbol{z}_i=\{z_{i,j}\}|_{j=1}^{|\boldsymbol{x}_i|}$ denotes mask values for each token in $i$th evidence.
We do not consider the sentence representation with $j=0$.
We then apply Hard Concrete reparameterization (HCR) trick~\citep{hard-concrete} to enforce the values approximate to discrete 0 or 1,
while keeping continuous and differential for learning mask vectors.
\begin{equation}
    \label{eq:token_rationales}
        \begin{split}
        \boldsymbol{z}_i = \boldsymbol{z}_i^{0} \odot \cdots \odot &\boldsymbol{z}_i^{L}, \quad \boldsymbol{pt}_i = \boldsymbol{pt}_i^{0} \odot \cdots \odot \boldsymbol{pt}_i^{L},\\
        (\boldsymbol{z}_{i}^{l}, \boldsymbol{pt}_{i}^{l}) &= \text{HCR}(g_t(\boldsymbol{h}^{l}_{i,j}|_{j=1}^{|\boldsymbol{x}_i|})),
        \end{split}
\end{equation}
where $\odot$ denotes Hadamard product, $pt_{i,j}|_{j=1}^{|\boldsymbol{x}_i|}\in \boldsymbol{pt}_i$ denotes the importance score of $j$th token in $i$th evidence.

For the sentence-level explainer,
we train a different interpreter network $g_s(\cdot)$ to predict a binary sentence mask vector $\boldsymbol{m}\in\mathbb{R}^n$ based on sentence representation to indicate the absence of sentence,
\begin{equation}
    \label{eq:sentence_rationale}
        \begin{split}
        \boldsymbol{m} = \boldsymbol{m}^{0} \odot \cdots \odot &\boldsymbol{m}^{L} ,\quad \boldsymbol{ps} = \boldsymbol{ps}^{0} \odot \cdots \odot \boldsymbol{ps}^{L}, \\
         (\boldsymbol{m}^{l},\boldsymbol{ps}^l) &= \text{HCR}(g_s(\tilde{\boldsymbol{h}}_{i,0}^{l}|_{i=0}^n)).
        \end{split}
\end{equation}

The multi-granular rationales are selected by multiplying the two mask vectors with the input\footnote{To ensure that the information of mask-out input is not propagated into the model inference, we employ an optional operation that masks the token and sentence representation in each layer of Transformer-XH with the two mask vectors.}, 
where token rationales $\boldsymbol{r}=\{\boldsymbol{r}_i\}|_{i=0}^n$ with $\boldsymbol{r}_i= \boldsymbol{x}_i\odot \boldsymbol{z}_i$ and sentence rationales $R_s=(X_s=X \odot \boldsymbol{m}, A_s=A \odot \boldsymbol{m}^\top \boldsymbol{m})$.
Therefore, the perturbed graph can be derived by intersecting the two subsets of granularity rationales, i.e., $G_R=(\{\boldsymbol{x_i}\cap \boldsymbol{r}_i\mid \boldsymbol{x_i} \in X_s\}, A_s)$.
Meanwhile,  
to ensure that only extracted rationale would be used for veracity prediction, 
we further intervene in the dynamic routing between the evidence capsule and the class capsule in the capsule network for succinct aggregation by multiplying the sentence mask vector with the coupling coefficients.

\subsection{Properties}
\label{subsec.property}
\paragraph{Fidelity} 
Fidelity guarantees that the model veracity is maintained after perturbing the input,
which measures the sufficiency for \textit{faithfulness} of multi-granular rationales~\citep{jiang-etal-2021-alignment-hardconcrete}.
To ensure the faithfulness of rationales,
We re-feed the original graph $G$ and perturbed graph $G_R$ into the veracity model $f(\cdot)$ to generate the prediction logits respectively.
Then we define the Euclidean distance between these two logits as fidelity loss,
\begin{equation}
    \label{eq:fidelity}
    \mathcal{L}_{F} = \Vert f(G) - f(G_R) \Vert_2.
\end{equation}

\paragraph{Consistency}
According to Definition~\ref{def:non-redundant},
we derive the decisive token rationale via improving the \textit{consistency} between the two single-granular explainers,
which ensures that almost all token rationales come from sentence rationales rather than from noise sentences.
We thus introduce the symmetric Jensen-Shannon Divergence to regularize the consistency between the importance score of two mask vectors,
\begin{equation}\small
    \label{eq:consistency}
    \begin{split}
    \mathcal{L}_{C} = &\frac{1}{2}\text{KL}(P(\boldsymbol{z})||\frac{P(\boldsymbol{z})+P(\boldsymbol{m})}{2}) \\
    &+\frac{1}{2}\text{KL}(P(\boldsymbol{m})||\frac{P(\boldsymbol{z})+P(\boldsymbol{m})}{2}),
    \end{split}
\end{equation}
where $P(\boldsymbol{z} )=\text{softmax}_i(\sum_{j=1}^{|\boldsymbol{x}_i|}pt_{i,j})$, $P(\boldsymbol{m})=\text{softmax}_i(ps_{i})$, and $\text{KL}(\cdot||\cdot)$ denotes the Kullback-Leibler divergence.
Clearly, the \textit{consistency} property tends to have the mutual effect that informative rationale on one side would help the other side.

\paragraph{Salience}
Unsupervised paradigm may be impracticable to extract high-quality multi-granular rationales. 
We thus utilize the predefined salience score as a signal to guide the rationale extraction as prior works.
For sentence rationale extraction,
following~\citet{informationbottleneck}, 
we adopt the rationale label as guidance by formulating it as a multi-label classification problem using cross entropy (CE) loss,
\begin{equation}
    \label{eq:salience_s}
\mathcal{L}_{SS} = \text{CE}(\boldsymbol{m}, \boldsymbol{E}),
\end{equation}
where $\boldsymbol{E}=\{E_i\in\{0,1\}\}|_{i=0}^n$ denotes whether the sentence is annotated rationale by humans.

Due to the expensive cost of gathering human rationale labels with fine-grained,
for token rationale extraction,
we construct the pseudo label $\boldsymbol{S}=\{\boldsymbol{s}_i\}|_{i=0}^n$ for each token in each piece of evidence via the technique of layered integrated gradient \citep{layer-integrated-gradient} provided by the Captum \citep{captum},
where $\boldsymbol{s}_i=\{s_{i,j}\in[-1,1]\}|_{j=0}^{|\boldsymbol{x}_i|}$.
Then the KL divergence is employed to regularize the token rationale extraction,
\begin{equation}
    \begin{split}
        \mathcal{L}_{ST} &= \sum_{i=0}^n \text{KL}(P(\boldsymbol{z}_i)||\hat{\boldsymbol{s}}_i),\\
    \end{split}
\end{equation}
where $P(\boldsymbol{z}_i) = \text{softmax}_j(pt_{i,j})$ denotes the importance score of tokens over the $i$th evidence, $\hat{\boldsymbol{s}}_i = \text{softmax}_j(s_{i,j})$.
In addition,
to regularize the \textit{compactness} of token rationale~\citep{jiang-etal-2021-alignment-hardconcrete},
we minimize the number of non-zeros predicted by the token-level explainer via minimizing the $\mathcal{L}_0$ norm with expectation ~\citep{decao2020}.
\begin{equation}
    \mathcal{L}_0 = \sum_{i}^n\sum_{j}^{|\boldsymbol{x}_i|}pt_{i,j},
\end{equation}

\subsection{Optimization}
The optimization objective is minimizing the following loss function $\mathcal{L}$,
\begin{equation}
\begin{split}
        \mathcal{L}  = \lambda_1 \mathcal{L}_{F} + \lambda_2 \mathcal{L}_{C} + \lambda_3 \mathcal{L}_{SS} + \lambda_4 \mathcal{L}_{ST} + \lambda_5 \mathcal{L}_{0},
\end{split}
\end{equation}
where $\lambda_{1-5}$ are hyperparameters standing for loss weights.

During training, we freeze the parameters of the pretrained veracity prediction model $f(\cdot)$ and only optimize the explainer parameters (i.e., $g_t(\cdot)$ and $g_s(\cdot)$) by minimizing $\mathcal{L}$. 
In the inference stage, the value of $z_{i,j}$ and $m_i$ are determined by $\mathbf{1}(pt_{ij} > \alpha)$ and $\mathbf{1}(ps_i > \alpha)$, respectively, 
where $\alpha$ is the threshold of rationales, 
$\mathbf{1}(\cdot)$ is the indicator function.

\section{Experiments}
\paragraph{Datasets}
We perform experiments on three multi-hop fact verification datasets, including {H}o{V}er \citep{hover}, LIAR-PLUS \citep{liar-plus}, and PolitiHop \citep{politihop}. 
For HoVer, following~\citep{baleen}, the dataset is constructed with retrieved evidence, where each claim is associated with 5 pieces of evidence. 
For LIAR-PLUS and PolitiHop, we use the datasets provided in \citet{politihop} and restrict each claim associated with 10 and 5 pieces of evidence, respectively. 
All the datasets require multi-hop reasoning and consist of annotated sentence-level rationale and noise evidence.

\paragraph{Baselines}
Since no other works aimed at multi-granular rationale extraction,
we compare CURE with twelve single-granular rationale extraction methods as baselines, 
including eight intrinsic-based methods 
(i.e., Pipeline in ERASER \citep{eraser2020},
Information Bottleneck (IB) \citep{informationbottleneck},
Two-Sentence Selecting (TSS) \citep{glockner2020},
Learning from rationales (LR) \citep{learning-from-rationales} for sentence rationale extraction.
\citet{lei2016},
DeClarE \citep{popat-etal-2018-declare},
FRESH \citep{fresh},
V\textsc{mask} \citep{vmask2020} for token rationale extraction.)
and four post hoc methods
(i.e., LIME \citep{lime}, 
SHAP \citep{shap_nips},
Layer Integrated Gradient (L-I\textsc{nt}G\textsc{rad}) \citep{layer-integrated-gradient},
D\textsc{iff}M\textsc{ask} \citep{decao2020}).

\paragraph{Metrics}
Inspired by \citet{eraser2020},
we adopt the macro F1 and accuracy for verification prediction evaluation, and macro F1, Precision and Recall to measure the sentence-level agreement with human-annotated rationales.
We also report fidelity defined in Equation~\ref{eq:fidelity} as a metric of faithfulness for post hoc methods.
We propose a metric named Token Rationale Overlap rate to measure the overlap between token rationale with sentence Rationale (\textbf{TRO-R}) or Non-rationale (\textbf{TRO-N}).
It reflects the consistency between the two granular rationales\footnote{This metric should be considered together with the evaluation of claim verification to avoid spurious high consistency.},
\begin{equation}\small
    \label{eq:trr}
    \begin{split}
        &\text{TRO-R}:=\frac{1}{|X_s|}\sum_{\boldsymbol{x_i}\in X_s}\frac{|\boldsymbol{x_i}\cap \boldsymbol{r_i}|}{|\boldsymbol{x_i}|}, \\
        &\text{TRO-N}:=\frac{1}{|X_{ns}|}\sum_{\boldsymbol{x_i}'\in X_{ns}}\frac{|\boldsymbol{x_i}'\cap \boldsymbol{r_i}|}{|\boldsymbol{x_i}'|}, \\
        &\text{Consistency}:= 1-\frac{\text{TRO-N}}{\text{TRO-R}}
    \end{split}
\end{equation}
where $X_{ns} = X\setminus X_s$ denotes the complement subset of $X_s$.

\paragraph{Implementation Details}
Our Veracity Prediction model adopts the pretrained RoBERTa~\citep{DBLP:journals/corr/abs-1907-11692} base model to initialize the Transformer components and three hop steps are used (i.e., $L=3$). 
The maximum number of input tokens to RoBERTa is 130 and the dimension of class capsule $d_c$ is 10. 
The pretrained model has 80.03\%, 83.14\%, and 71.63\% on label accuracy of claim verification on HoVer, LIAR-PLUS, and PolitiHop, respectively. 

\section{Results and Discussion}
\label{sec:result_and_discussion}
\subsection{Quantitative Analysis}
\label{Quantitative analysis}

\begin{table*}[t]
    \centering
    \resizebox{!}{6.8cm}{
    \begin{tabular}{llcccccc}
    \toprule[2pt]
    \multirow{2.5}{*}{\makebox[0.15\textwidth][l]{\textbf{Dataset}}}    & \multirow{2.5}{*}{\makebox[0.2\textwidth][l]{\textbf{Model}}} & \multicolumn{3}{c}{\textbf{Claim Verification}}          & \multicolumn{3}{c}{\textbf{Rationale Extraction}}            \\
    \cmidrule(lr){3-5}\cmidrule(lr){6-8}
                                         &                                 & \makebox[0.15\textwidth][c]{\textbf{Acc.}} & \makebox[0.15\textwidth][c]{\textbf{F1}}   & \makebox[0.15\textwidth][c]{\textbf{Fidelity$\downarrow$}} & \makebox[0.15\textwidth][c]{\textbf{TRO-R${\uparrow}$}} & \makebox[0.15\textwidth][c]{\textbf{TRO-N$\downarrow$}} & \makebox[0.15\textwidth][c]{\textbf{Consistency$\uparrow$}} \\\midrule[1pt]
    \multirow{13}{*}{\textbf{LIAR-PLUS}} & \textbf{\citet{lei2016}}     & 0.5681 & 0.5442                    & -/-                     & 0.0042            & 0.0031             & 0.2619               \\
                                         & \textbf{DeClarE}               & 0.4773  & 0.2154                   & -/-                     & 0.2972            & 0.2981             & -0.0030              \\
                                         & \textbf{FRESH}                 & 0.4345  & 0.4137                   & -/-                     & 0.3810            & 0.2697             & 0.2921               \\
                                         & \textbf{V\textsc{mask}}                  & 0.8262 & 0.8146                    & -/-                     & 0.3542            & 0.3614             & -0.0203              \\
                                         & \textbf{LIME}          & 0.3061          & 0.1562                   &0.8422                  & 0.1077            & 0.0558             & 0.4819               \\
                                         & \textbf{SHAP}        & 0.7639           & 0.7531                    & 1.6401                    & 0.8572            & 0.8477             & 0.0111               \\
                                         & \textbf{L-I\textsc{nt}G\textsc{rad}}        & 0.7172          & 0.6984                   &0.5889                      & 0.4905            & 0.4802             & 0.0210               \\
                                         & \textbf{D\textsc{iff}M\textsc{ask}}          & 0.5850      & 0.4803                   & 4.2244                    & 0.3352            & 0.3308             & 0.0131               \\\cdashline{2-8}
                                         & \textbf{CURE*}       & \textbf{0.8210}             & \textbf{0.8078} & \textbf{0.2675}            & 0.4031   & 0.1491    & \textbf{0.6301}      \\\cdashline{2-8}
                                         & \textbf{CURE}       & 0.8210            & 0.8078  & 0.2675            & 0.3287   & 0.1577    & \textbf{0.5202}      \\
                                         & \textbf{CURE} -\emph{C}           & 0.8132     & 0.8028                    & 0.2642                     & 0.2488            & 0.1831             & 0.2641               \\
                                         & \textbf{CURE} -\emph{SS}             & 0.7704     & 0.7502                    & 0.3101                     & 0.1114            & 0.0672             & 0.3968               \\
                                         & \textbf{CURE} -\emph{ST}             & 0.8171      & 0.8069                    & \textbf{0.2541}                     & 0.3825            & 0.2389             & 0.3754               \\\midrule[1pt]
    \multirow{13}{*}{\textbf{HoVer}} & \textbf{\citet{lei2016}}                    & 0.5015 & 0.3410                   & -/-                     & 0.0015            & 0.0014             & 0.0667               \\   
                                         & \textbf{DeClarE}       & 0.5083          & 0.5076                   &-/-                      & 0.5307            & 0.4212             & 0.2063               \\
                                         & \textbf{FRESH}        & 0.6028          & 0.6014                    &-/-                      & 0.3250            & 0.5646             & -0.7372              \\
                                         & \textbf{V\textsc{mask}}                 & 0.7438   & 0.7369                   &-/-                      & 0.5443            & 0.3418             & 0.3720               \\
                                         & \textbf{LIME}          & 0.5000         & 0.3333                    & 0.8356                        & 0.0982            & 0.0287             & 0.7077               \\
                                         & \textbf{SHAP}          & 0.5983           & 0.5818                  & 2.6125                      & 0.7402            & 0.6758             & 0.0870               \\
                                         & \textbf{L-I\textsc{nt}G\textsc{rad}}          & 0.5003         & 0.5386                  & 0.7058                     & 0.4514            & 0.3651             & 0.1912               \\
                                         &\textbf{D\textsc{iff}M\textsc{ask}}        & 0.7153       & 0.7130                   & 1.1632                      & 0.1781            & 0.0628             & 0.6474               \\\cdashline{2-8}
                                         & \textbf{CURE*}   & \textbf{0.7698}                & \textbf{0.7689}  & \textbf{0.2287}            & 0.6993   & 0.0334    & \textbf{0.9522}      \\\cdashline{2-8}
                                         & \textbf{CURE}       &0.7698            & 0.7689 &   \textbf{0.2287}            & 0.6986   & 0.1777    & 0.7456      \\       
                                         & \textbf{CURE} -\emph{C}            & 0.7585      & 0.7561                   &0.2405                      & 0.5837            & 0.1529             & 0.7381               \\
                                         & \textbf{CURE} -\emph{SS}              & 0.7298      & 0.7297                    &0.3469                      & 0.3646            & 0.1831             & 0.4978               \\
                                         & \textbf{CURE} -\emph{ST}              & 0.7683     & 0.7672                    &0.2330                      & 0.6620             & 0.1671             & \textbf{0.7476}               \\\midrule[1pt]
    \multirow{13}{*}{\textbf{PolitiHop}}  & \textbf{\citet{lei2016}}                 & 0.5674    & 0.3691                   &-/-                & 0.0047            & 0.0040             & 0.1489               \\
                                         & \textbf{DeClarE}      & 0.6950           & 0.2734                   & -/-                     & 0.4994            & 0.4468             & 0.1053               \\
                                         & \textbf{FRESH}     & 0.6170             & 0.4435                    &-/-                      & 0.4152            & 0.3527             & 0.1505               \\
                                         & \textbf{V\textsc{mask}}                & 0.7234    & 0.5580                   &-/-                      & 0.4142            & 0.4111             & 0.0075               \\
                                         & \textbf{LIME}       & 0.6950              & 0.2734                  &0.8041                      & 0.0528            & 0.0525             & 0.0057               \\
                                         & \textbf{SHAP}    & 0.5957                & 0.4071                   &2.2659                      & 0.7721            & 0.7834             & -0.0146              \\
                                         & \textbf{L-I\textsc{nt}G\textsc{rad}}          & 0.6950        & 0.2734                  & 0.6580                     & 0.5108            & 0.5097             & 0.0022               \\
                                         & \textbf{D\textsc{iff}M\textsc{ask}}         & 0.6738      & 0.4471                   & 2.4533                     & 0.3146            & 0.3104             & 0.0134               \\\cdashline{2-8}
                                         & \textbf{CURE*}       & \textbf{0.6950}            & \textbf{0.3236}  & \textbf{0.3204}            & 0.6391   & 0.0828    & \textbf{0.8704}      \\\cdashline{2-8}
                                         & \textbf{CURE}        & 0.6950           & 0.3236  & 0.3204            & 0.5515   & 0.3258    & \textbf{0.4092}      \\
                                         & \textbf{CURE} -\emph{C}        & 0.6525         & 0.3553                   &0.2984                      & 0.6168            & 0.3864             & 0.3735               \\
                                         & \textbf{CURE} -\emph{SS}            & 0.6950        & 0.4214                    &\textbf{0.2563}                      & 0.3381            & 0.2881             & 0.1479               \\
                                         & \textbf{CURE} -\emph{ST}            & 0.6809        & 0.2951                   & 0.3372                     & 0.5643            & 0.3455             & 0.3877               \\\bottomrule[2pt]                      
    \end{tabular}}
    \caption{Evaluation results of multi-granular rationale across three datasets.
    \textbf{CURE*} denotes the results using predicted token rationale and predicted sentence rationale,
    \textbf{CURE} denotes the results using predicted token rationale and annotated sentence rationale.
    $\uparrow$ means the larger value is better.
    $-C$, $-SS$ and $-ST$ denote the constraint removal of \textit{Consistency}, \textit{Salience-Sentence} and \textit{Salience-Token}, respectively. Our main results are marked in bold.
    }
    \label{tab:token_results}
\end{table*}

\begin{table*}[t]
    \centering
    \scriptsize
    \begin{tabular}{llccccc}
    \toprule[2pt]
    \multirow{2.5}{*}{\makebox[0.1\textwidth][l]{\textbf{Dataset}}}   & \multirow{2.5}{*}{\makebox[0.1\textwidth][l]{\textbf{Model}}} & \multicolumn{2}{c}{\textbf{Claim Verification}} & \multicolumn{3}{c}{\textbf{Sentence Rationale}} \\
    \cmidrule(lr){3-4}\cmidrule(lr){5-7}
                                        &                                 & \makebox[0.08\textwidth][c]{\textbf{Acc.}}  & \makebox[0.08\textwidth][c]{\textbf{F1}}  & \makebox[0.08\textwidth][c]{\textbf{F1}}   & \makebox[0.08\textwidth][c]{\textbf{Precision}}  & \makebox[0.08\textwidth][c]{\textbf{Recall}}  \\\midrule[1pt]
    \multirow{5}{*}{\textbf{LIAR-PLUS}} & \textbf{Pipeline}        & 0.5811         & 0.5393                   & 0.6677            & 0.7450           & 0.6564           \\
                                        & \textbf{IB}       & 0.6252                & 0.6048                   & 0.3777            & 0.3927           & 0.3967           \\
                                        & \textbf{TSS}     & 0.6239               & 0.6172                     & 0.4324            & 0.6349           & 0.3469           \\
                                        & \textbf{LR}     & 0.7652                & 0.7519                     & 0.6242            & 0.6776           & 0.6381           \\\cdashline{2-7}
                                        & \textbf{CURE}    & \textbf{0.8210}       & \textbf{0.8078} & \textbf{0.6789} & \textbf{0.8072} & \textbf{0.6329}         \\\midrule[1pt]
    \multirow{5}{*}{\textbf{HoVer}}     & \textbf{Pipeline}         & 0.6255       & 0.6244                    & \textbf{0.9427}            & 0.9028           & 0.9900           \\
                                        & \textbf{IB}      & 0.5678               & 0.5674                     & 0.6236            & 0.7018           & 0.5783           \\
                                        & \textbf{TSS}       & 0.5368             & 0.5111                     & 0.6883            & 0.9026           & 0.5755           \\
                                        & \textbf{LR}     & 0.5110                 & 0.4050                    & 0.9419            & 0.9029           & \textbf{0.9988}           \\\cdashline{2-7}
                                        & \textbf{CURE}       & \textbf{0.7698}    & \textbf{0.7689} & 0.9376 & \textbf{0.9045} & 0.9877   \\\midrule[1pt]
    \multirow{5}{*}{\textbf{PolitiHop}} & \textbf{Pipeline}        & 0.6596       & 0.4173                     & 0.6390            & 0.5986           & 0.8234           \\
                                        & \textbf{IB}       & 0.6879                & \textbf{0.5489}                   & 0.4180            & 0.5106           & 0.3902           \\
                                        & \textbf{TSS}       & 0.6525             & 0.4334                     & 0.4272            & 0.5177           & 0.4044           \\
                                        & \textbf{LR}       & \textbf{0.7021}               & 0.4712                    & 0.5699            & 0.5674           & 0.6657           \\\cdashline{2-7}
                                        & \textbf{CURE}   & 0.6950       & 0.3459  & \textbf{0.6947} & \textbf{0.6584} & \textbf{0.8403}     \\\bottomrule[2pt]      
    \end{tabular}
    \caption{
    Evaluation of claim verification and sentence rationale extraction across three datasets. The best results are marked in bold.}
    \label{tab:sent_results}
    \end{table*}

    \begin{table}[t]
        \centering
        \resizebox{7cm}{!}{\begin{tabular}{lcccc}\toprule[2pt]
        \textbf{Model}     & \textbf{Spearman} & \textbf{F1}     & \textbf{Precision} & \textbf{Recall} \\\midrule[1pt]
        \textbf{LIME}      & 0.1695            & 0.5422          & 0.6564             & 0.5459          \\
        \textbf{SHAP}      & 0.0305            & 0.3636          & 0.5138             & 0.5170          \\
        \textbf{L-I\textsc{nt}G\textsc{rad}} & 0.0776            & 0.5108          & 0.5314             & 0.5479          \\
        \textbf{V\textsc{mask}}     & 0.1177            & 0.5247          & 0.5473             & 0.5732          \\
        \textbf{CURE}      & \textbf{0.4293}   & \textbf{0.6747} & \textbf{0.6739}    & \textbf{0.7650}  \\\bottomrule[2pt]
        \end{tabular}}
        \caption{
        Evaluation of token rationale extraction on the HoVer dataset based on our re-annotation.
        The best results are marked in bold.}
        \label{tab:human-agree}
        \end{table}

\paragraph{Main results}
Table~\ref{tab:token_results} presents the results from our CURE against the baselines for claim verification and rationale extraction.
We report our main evaluation of the multi-granular rationale extraction on -CURE*.
Moreover, since the baselines cannot extract the two granular rationales simultaneously,
for a fair comparison, we also report the evaluation using the sentence rationale annotated by humans instead of the predicted sentence rationale to compute the TRO-R and TRO-N. 
We can observe that:
(I) CURE is quite \textbf{faithful} with the lowest fidelity value across all three datasets, surpassing all other baselines.
This result is in accordance with~\citet{jiang-etal-2021-alignment-hardconcrete} that the Euclidean distance between the logits constrains the explainer to provide more faithful explanations.
(II) CURE is capable of extracting \textbf{consistent} multi-granular rationales with the highest consistency score,
which indicates the importance of the differential between true evidence and noise evidence for the token rationale extraction.
This is significantly reflected in the CURE*. 
In contrast, all baselines are unable to induce consistent rationales with huge gaps towards our CURE, even though some baselines achieve better performance on single TRO-R or TRO-N (e.g., SHAP on TRO-R and LIME on TRO-N).
(III) On claim verification, our CURE outperforms the post hoc methods, while slightly lower compared with intrinsic methods.
We conjecture that the information leakage caused by soft selection may improve the performance of these models.
(IV) Beyond relative performance against baselines,
we conduct control experiments in the ablation study to explore the effectiveness of \textbf{diagnostic property}. 
With the removal of different properties individually,
we observe the reduced performance in the extracted rationales, both in fidelity and consistency.
The most significant property is Salience-Sentence,  
this can be due to that explainer is susceptible to over-fitting and yields task-irrelevant token explanations from noise sentences when lacking prior knowledge about the data.
The second key property is Consistency,
there are varying decreases in both fidelity and consistency throughout the three datasets,
particularly for LIAR-PLUS, which requires more complex rationales for reasoning over multiple evidence compared with the other two datasets.
We reasonably presume the synergy of the two granular explainers by constraining the extraction of \textit{right token} from \textit{right sentence}~\citep{gupta-right}.
Moreover, we note a minor decrease for claim verification when removing the Salience-Token,
showing that the retained task-relevant tokens directed by the salience score can help to boost the performance of veracity prediction.

\paragraph{Plausibility} 
As shown in Table~\ref{tab:sent_results} and~\ref{tab:human-agree},
we further conduct the experiments to explore how well the extracted rationales agrees with human annotation~\citep{jacovi-goldberg-2020} compared to classical single-granular rationale methods.

For \textbf{sentence rationale}, surprisingly,
we find that our CURE still outperforms the most baselines on claim verification and rationale extraction.
We reasonably posit that the high quality \textit{right token} is useful for extracting \textit{right sentence} rationale in turn.
To further validate the quality of \textbf{token rationale} extraction,
we ask 3 annotators with NLP backgrounds to re-annotate 150 fine-grained samples from the development set of the HoVer dataset to obtain the rationale label at the token level.
Our annotators achieve 0.6807 on Krippendorff's $\alpha$ \citep{krippendorff2011computing} and retain 20\% tokens annotated as rationales.
We measure the agreement between the predicted token rationale and human annotated rationale with the \emph{Spearman's correlation}, \emph{macro F1}, \emph{Precision}, and \emph{Recall}.
As shown in Table~\ref{tab:human-agree},
our CURE is far more promising that outperforms the baselines with a huge gap on all evaluation metrics.
It clearly indicates the necessity of consistency between multi-granular rationales for explaining multi-hop fact verification.

\subsection{Manual Evaluation}
\label{subsec.human_evaluation}
Inspired by \citet{NEURIPS2020_zhou} and \citet{yan2022hierarchical},
we provide a manual evaluation of the token rationales (contained in the sentence rationale rather than the whole sentences) extracted by CURE, compared to D\textsc{iff}M\textsc{ask} \citep{decao2020} and V\textsc{mask} \citep{vmask2020}.
We randomly select 50 samples and ask three annotators with NLP backgrounds to score these rationales in a likert scale of 1 to 5 according to three different criteria:
(\uppercase\expandafter{\romannumeral1}) \textbf{Correctness}, which measures what extent users can approach ground-truth label given the predicted token rationales;
(\uppercase\expandafter{\romannumeral2}) \textbf{Faithfulness}, which measures what extent users can approach the model predicted label given the predicted token rationales;
(\uppercase\expandafter{\romannumeral3}) \textbf{Non-redundancy}, which measures what extent the predicted token rationales do not contain redundant and irrelevant words.

\begin{figure}[t]
    \centering
    \includegraphics[width=7.5cm]{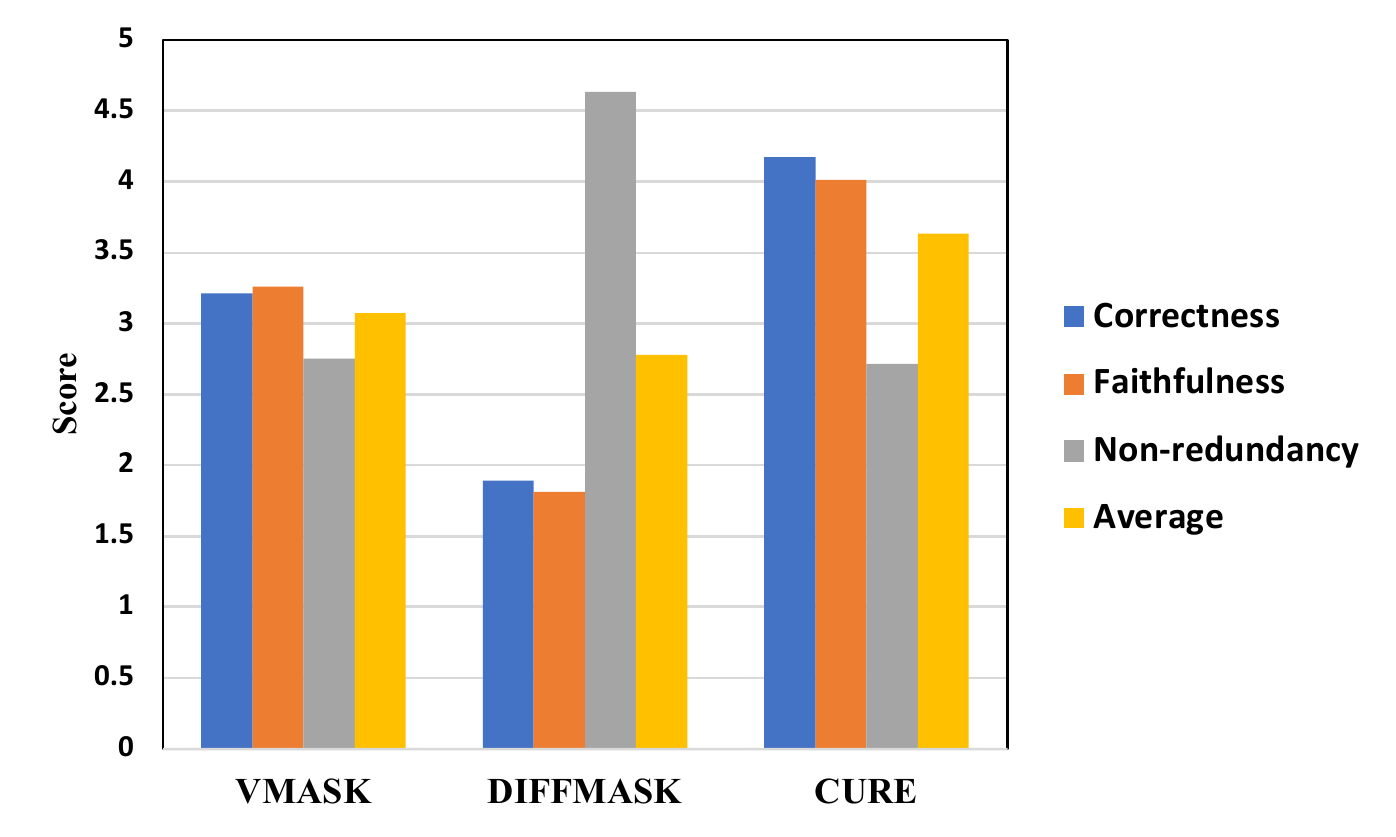}
    \caption{Human evaluation results in a likert scale of 1 to 5, where 1 means strongly disagree and 5 means strongly agree. \emph{Average} denotes the average score of three criteria. The inner-rater agreement measured by Krippendorff's $\alpha$ is 0.88.}
    \label{fig:human_evaluation}
\end{figure}

The human evaluation results are shown in Figure~\ref{fig:human_evaluation}. We can observe that CURE achieves the best results on correctness and faithfulness. 
Although D\textsc{iff}M\textsc{ask} performs particularly well on non-redundancy, the correctness and faithfulness of the generated rationales are far worse than those of the other two models, 
indicating the low quality of its rationales. In fact, D\textsc{iff}M\textsc{ask} excels at masking almost all tokens due to the only constraint of $L_0$ loss. 
Considering the mutual constraints between non-redundancy and the other two criteria, we calculate the average scores of three criteria for each method. 
CURE still outperforms on average score, which demonstrates the high quality of the token rationales generated by our method.

\begin{figure}[t]
    \centering
    \includegraphics[width=7.4cm]{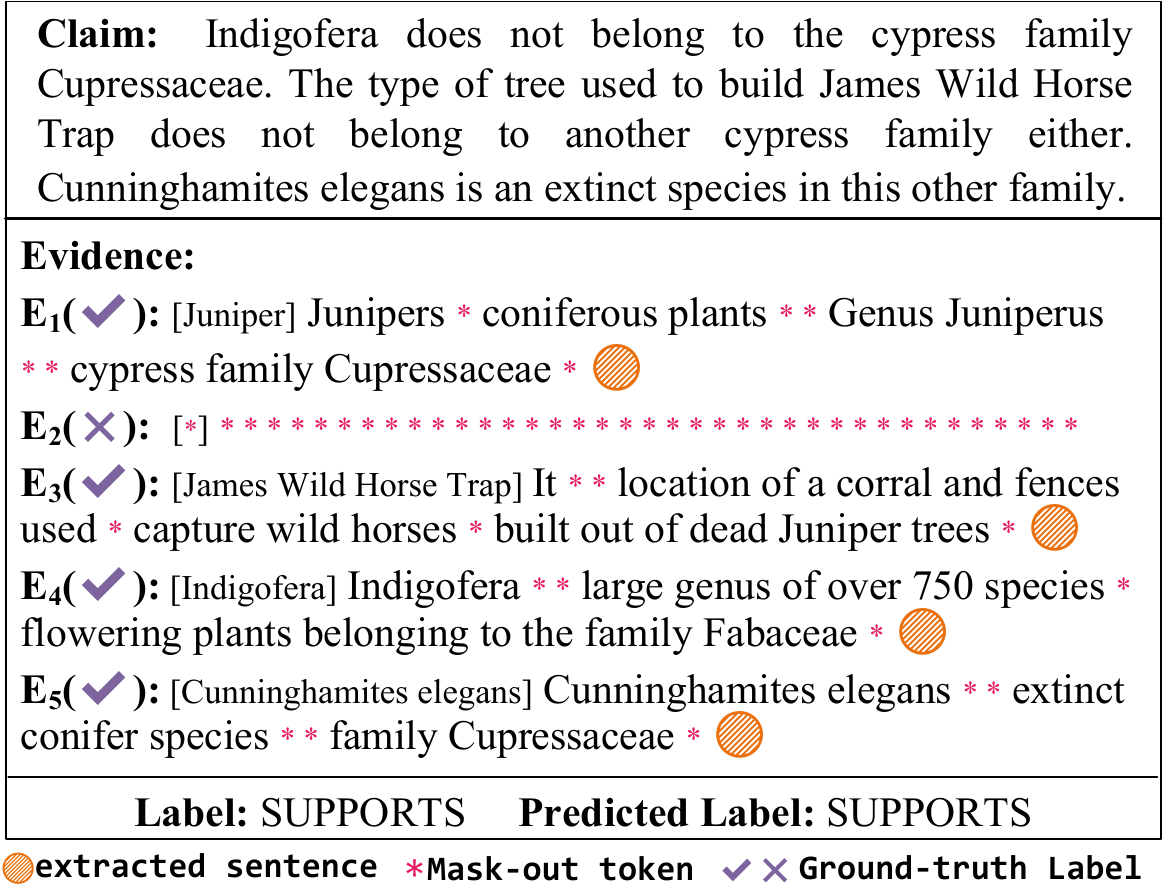}
    \caption{Rationales extracted by our method on the HoVer dataset. 
    All the tokens except \textcolor[RGB]{233,51,127}{*} denote the token rationales predicted by our CURE.}
    \label{fig:case_study_hover}
\end{figure}

\subsection{Rationale Examples}
Figure~\ref{fig:case_study_hover} presents an intuitive example with rationales generated by our CURE from the HoVer dataset.
We can observe that our CURE correctly predicts the sentence rationales while entirely removing the noise sentence $E2$.
Meanwhile,
The corresponding retained token rationales contain information that is not only important for veracity prediction,
but also appears the non-redundancy of the token rationales by ignoring the redundant tokens.
Moreover,
the retained tokens show strong consistency towards the extracted sentence rationales.
It is worth noting that our CURE is prone to retaining the \textit{title} of the document as the key cue for linking multiple pieces of evidence.

\section{Related Work}
A growing interest in interpretability has led to a flurry of approaches in trying to reveal the reasoning behavior behind the multi-hop fact verification task.
A well-studied way is to use the attention weights as attribution score to indicate the importance of a token or sentence, such as self-attention~\citep{popat-etal-2018-declare} or co-attention~\citep{defend-kdd, xfact-www19, wu-etal-2020-dtca, wu-etal-2021-unified}.
While this method is incapable to guarantee the inattention of low-score features, drawing criticism recently~\citep{DBLP:conf/emnlp/WiegreffeP19,DBLP:conf/acl/MeisterLAC20}.
Another line of research focuses on perturbation-based methods.
These methods explore a built-in explainer to generate the rationale by masking the unimportant language features~\citep{generating-fact2020, informationbottleneck, glockner2020, pubhealth-2020, predictagain2021, claimdissector2022}.
This way generally employs the \textit{extract-then-predict} paradigm,
while \citet{yu2021understanding} reveals an issue of model interlocking in such a cooperative rationalization paradigm.

Recently, a few studies explore the post hoc paradigm for explanation extraction by detaching the explainer and the task model.
With the parameters of the task model frozen,
they focus on the external explainer to retain the key cue in input as the rationales to indicate features the task model relies on \citep{decao2020,si2022exploring,atanasov2022,evarm2022}.
Our work falls under the scope of the post hoc paradigm,
different from the prior works that only consider the single-granular rationale, 
we for the first time propose a novel paradigm to yield indicative token rationales by regularizing the multi-granular rationale extraction.

\section{Conclusion}
In this paper, we propose a novel multi-granular rationale extraction framework for explainable multi-hop fact verification. We jointly model token-level and sentence-level rationale extraction by incorporating three diagnostic properties as additional constraints to generate faithful and consistent multi-granular rationales. The results on three multi-hop fact verification datasets illustrate the effectiveness of our method. In the future, we will explore how to generate counterfactual explanations.

\section*{Limitations}
A limitation of our work is that we employ the supervised paradigm because of the difficulty to satisfy our expectations about the rationales. 
We need the labels of sentence-level rationales as guidance to obtain better classification performance and high-quality rationales, 
which may be difficult to extend our method into the scenarios with few annotations (i.e., semi-supervised or unsupervised). 
In addition, the $L_0$ loss regularization overemphasizes the sparsity, which can damage the performance on claim verification and make the model sensitive to hyperparameters.

\bibliography{mybib}

\begin{thebibliography}{49}
\expandafter\ifx\csname natexlab\endcsname\relax\def\natexlab#1{#1}\fi

\bibitem[{Alhindi et~al.(2018)Alhindi, Petridis, and Muresan}]{liar-plus}
Tariq Alhindi, Savvas Petridis, and Smaranda Muresan. 2018.
\newblock \href {https://doi.org/10.18653/v1/W18-5513} {Where is your evidence:
  Improving fact-checking by justification modeling}.
\newblock In \emph{Proceedings of the First Workshop on Fact Extraction and
  {VER}ification ({FEVER})}, pages 85--90.

\bibitem[{Atanasova et~al.(2020)Atanasova, Simonsen, Lioma, and
  Augenstein}]{generating-fact2020}
Pepa Atanasova, Jakob~Grue Simonsen, Christina Lioma, and Isabelle Augenstein.
  2020.
\newblock \href {https://doi.org/10.18653/v1/2020.acl-main.656} {Generating
  fact checking explanations}.
\newblock In \emph{Proceedings of the Annual Meeting of the Association for
  Computational Linguistics}, pages 7352--7364.

\bibitem[{Atanasova et~al.(2022)Atanasova, Simonsen, Lioma, and
  Augenstein}]{atanasov2022}
Pepa Atanasova, Jakob~Grue Simonsen, Christina Lioma, and Isabelle Augenstein.
  2022.
\newblock \href {https://ojs.aaai.org/index.php/AAAI/article/view/21287}
  {Diagnostics-guided explanation generation}.
\newblock In \emph{Proceedings of the AAAI Conference on Artificial
  Intelligence}, pages 10445--10453.

\bibitem[{Carton et~al.(2022)Carton, Kanoria, and
  Tan}]{learning-from-rationales}
Samuel Carton, Surya Kanoria, and Chenhao Tan. 2022.
\newblock \href {https://doi.org/10.18653/v1/2022.findings-acl.86} {What to
  learn, and how: {T}oward effective learning from rationales}.
\newblock In \emph{Proceedings of Findings of the Association for Computational
  Linguistics: ACL}, pages 1075--1088.

\bibitem[{Chen and Ji(2020)}]{vmask2020}
Hanjie Chen and Yangfeng Ji. 2020.
\newblock \href {https://doi.org/10.18653/v1/2020.emnlp-main.347} {Learning
  variational word masks to improve the interpretability of neural text
  classifiers}.
\newblock In \emph{Proceedings of the Conference on Empirical Methods in
  Natural Language Processing}, pages 4236--4251.

\bibitem[{De~Cao et~al.(2020)De~Cao, Schlichtkrull, Aziz, and
  Titov}]{decao2020}
Nicola De~Cao, Michael~Sejr Schlichtkrull, Wilker Aziz, and Ivan Titov. 2020.
\newblock \href {https://doi.org/10.18653/v1/2020.emnlp-main.262} {How do
  decisions emerge across layers in neural models? interpretation with
  differentiable masking}.
\newblock In \emph{Proceedings of the Conference on Empirical Methods in
  Natural Language Processing}, pages 3243--3255.

\bibitem[{DeYoung et~al.(2020)DeYoung, Jain, Rajani, Lehman, Xiong, Socher, and
  Wallace}]{eraser2020}
Jay DeYoung, Sarthak Jain, Nazneen~Fatema Rajani, Eric Lehman, Caiming Xiong,
  Richard Socher, and Byron~C. Wallace. 2020.
\newblock \href {https://doi.org/10.18653/v1/2020.acl-main.408} {{ERASER}: {A}
  benchmark to evaluate rationalized {NLP} models}.
\newblock In \emph{Proceedings of the Annual Meeting of the Association for
  Computational Linguistics}, pages 4443--4458.

\bibitem[{Fajcik et~al.(2022)Fajcik, Motlicek, and Smrz}]{claimdissector2022}
Martin Fajcik, Petr Motlicek, and Pavel Smrz. 2022.
\newblock \href {https://arxiv.org/abs/2207.14116} {Claim-dissector: An
  interpretable fact-checking system with joint re-ranking and veracity
  prediction}.
\newblock \emph{arXiv preprint}, arXiv:2207.14116.

\bibitem[{Feng et~al.(2018)Feng, Wallace, Grissom~II, Iyyer, Rodriguez, and
  Boyd-Graber}]{feng-etal-2018-pathologies}
Shi Feng, Eric Wallace, Alvin Grissom~II, Mohit Iyyer, Pedro Rodriguez, and
  Jordan Boyd-Graber. 2018.
\newblock \href {https://doi.org/10.18653/v1/D18-1407} {Pathologies of neural
  models make interpretations difficult}.
\newblock In \emph{Proceedings of the Conference on Empirical Methods in
  Natural Language Processing}, pages 3719--3728.

\bibitem[{Ge et~al.(2022)Ge, Hu, Ma, Wu, Chen, Liu, Zhang, Qin, and
  Zhang}]{evarm2022}
Ling Ge, ChunMing Hu, Guanghui Ma, Junshuang Wu, Junfan Chen, JiHong Liu, Hong
  Zhang, Wenyi Qin, and Richong Zhang. 2022.
\newblock \href {https://aclanthology.org/2022.coling-1.87} {{E}-{V}ar{M}:
  Enhanced variational word masks to improve the interpretability of text
  classification models}.
\newblock In \emph{Proceedings of the International Conference on Computational
  Linguistics}, pages 1036--1050.

\bibitem[{Glockner et~al.(2020)Glockner, Habernal, and Gurevych}]{glockner2020}
Max Glockner, Ivan Habernal, and Iryna Gurevych. 2020.
\newblock \href {https://doi.org/10.18653/v1/2020.findings-emnlp.97} {Why do
  you think that? exploring faithful sentence-level rationales without
  supervision}.
\newblock In \emph{Proceedings of Findings of the Association for Computational
  Linguistics: EMNLP}, pages 1080--1095.

\bibitem[{Gupta et~al.(2022)Gupta, Zhang, Vempala, He, Choji, and
  Srikumar}]{gupta-right}
Vivek Gupta, Shuo Zhang, Alakananda Vempala, Yujie He, Temma Choji, and Vivek
  Srikumar. 2022.
\newblock \href {https://doi.org/10.18653/v1/2022.acl-long.231} {Right for the
  right reason: Evidence extraction for trustworthy tabular reasoning}.
\newblock In \emph{Proceedings of the Annual Meeting of the Association for
  Computational Linguistics}, pages 3268--3283.

\bibitem[{Jacovi and Goldberg(2020)}]{jacovi-goldberg-2020}
Alon Jacovi and Yoav Goldberg. 2020.
\newblock \href {https://doi.org/10.18653/v1/2020.acl-main.386} {Towards
  faithfully interpretable {NLP} systems: How should we define and evaluate
  faithfulness?}
\newblock In \emph{Proceedings of the Annual Meeting of the Association for
  Computational Linguistics}, pages 4198--4205.

\bibitem[{Jain et~al.(2020)Jain, Wiegreffe, Pinter, and Wallace}]{fresh}
Sarthak Jain, Sarah Wiegreffe, Yuval Pinter, and Byron~C. Wallace. 2020.
\newblock \href {https://doi.org/10.18653/v1/2020.acl-main.409} {{L}earning to
  faithfully rationalize by construction}.
\newblock In \emph{Proceedings of the Annual Meeting of the Association for
  Computational Linguistics}, pages 4459--4473.

\bibitem[{Janizek et~al.(2021)Janizek, Sturmfels, and Lee}]{janizek2021}
Joseph~D. Janizek, Pascal Sturmfels, and Su-In Lee. 2021.
\newblock \href {https://www.jmlr.org/papers/volume22/20-1223/20-1223.pdf}
  {Explaining explanations: Axiomatic feature interactions for deep networks}.
\newblock \emph{Journal of Machine Learning Research}, pages 1--54.

\bibitem[{Jiang et~al.(2020)Jiang, Bordia, Zhong, Dognin, Singh, and
  Bansal}]{hover}
Yichen Jiang, Shikha Bordia, Zheng Zhong, Charles Dognin, Maneesh Singh, and
  Mohit Bansal. 2020.
\newblock \href {https://doi.org/10.18653/v1/2020.findings-emnlp.309}
  {{H}o{V}er: A dataset for many-hop fact extraction and claim verification}.
\newblock In \emph{Proceedings of Findings of the Association for Computational
  Linguistics: EMNLP}, pages 3441--3460.

\bibitem[{Jiang et~al.(2021)Jiang, Zhang, Yang, Zhao, and
  Liu}]{jiang-etal-2021-alignment-hardconcrete}
Zhongtao Jiang, Yuanzhe Zhang, Zhao Yang, Jun Zhao, and Kang Liu. 2021.
\newblock \href {https://doi.org/10.18653/v1/2021.acl-long.417} {Alignment
  rationale for natural language inference}.
\newblock In \emph{Proceedings of the Annual Meeting of the Association for
  Computational Linguistics and the International Joint Conference on Natural
  Language Processing}, pages 5372--5387.

\bibitem[{Khattab et~al.(2021)Khattab, Potts, and Zaharia}]{baleen}
Omar Khattab, Christopher Potts, and Matei Zaharia. 2021.
\newblock \href
  {https://proceedings.neurips.cc/paper/2021/file/e8b1cbd05f6e6a358a81dee52493dd06-Paper.pdf}
  {Baleen: Robust multi-hop reasoning at scale via condensed retrieval}.
\newblock In \emph{Advances in Neural Information Processing Systems},
  volume~34, pages 27670--27682.

\bibitem[{Kokhlikyan et~al.(2020)Kokhlikyan, Miglani, Martin, Wang, Alsallakh,
  Reynolds, Melnikov, Kliushkina, Araya, Yan, and
  Reblitz{-}Richardson}]{captum}
Narine Kokhlikyan, Vivek Miglani, Miguel Martin, Edward Wang, Bilal Alsallakh,
  Jonathan Reynolds, Alexander Melnikov, Natalia Kliushkina, Carlos Araya, Siqi
  Yan, and Orion Reblitz{-}Richardson. 2020.
\newblock \href {http://arxiv.org/abs/2009.07896} {Captum: {A} unified and
  generic model interpretability library for pytorch}.
\newblock \emph{CoRR}, abs/2009.07896.

\bibitem[{Kotonya and Toni(2020{\natexlab{a}})}]{survey2020}
Neema Kotonya and Francesca Toni. 2020{\natexlab{a}}.
\newblock \href {https://doi.org/10.18653/v1/2020.coling-main.474} {Explainable
  automated fact-checking: A survey}.
\newblock In \emph{Proceedings of the International Conference on Computational
  Linguistics}, pages 5430--5443.

\bibitem[{Kotonya and Toni(2020{\natexlab{b}})}]{pubhealth-2020}
Neema Kotonya and Francesca Toni. 2020{\natexlab{b}}.
\newblock \href {https://doi.org/10.18653/v1/2020.emnlp-main.623} {Explainable
  automated fact-checking for public health claims}.
\newblock In \emph{Proceedings of the Conference on Empirical Methods in
  Natural Language Processing}, pages 7740--7754.

\bibitem[{Krippendorff(2011)}]{krippendorff2011computing}
Klaus Krippendorff. 2011.
\newblock \href {https://repository.upenn.edu/asc_papers/43/} {Computing
  krippendorff's alpha-reliability}.

\bibitem[{Lei et~al.(2016)Lei, Barzilay, and Jaakkola}]{lei2016}
Tao Lei, Regina Barzilay, and Tommi Jaakkola. 2016.
\newblock \href {https://doi.org/10.18653/v1/D16-1011} {Rationalizing neural
  predictions}.
\newblock In \emph{Proceedings of the Conference on Empirical Methods in
  Natural Language Processing}, pages 107--117.

\bibitem[{Li et~al.(2016)Li, Monroe, and Jurafsky}]{DBLP:journals/corr/LiMJ16a}
Jiwei Li, Will Monroe, and Dan Jurafsky. 2016.
\newblock \href {http://arxiv.org/abs/1612.08220} {Understanding neural
  networks through representation erasure}.
\newblock \emph{CoRR}, abs/1612.08220.

\bibitem[{Liu et~al.(2019)Liu, Ott, Goyal, Du, Joshi, Chen, Levy, Lewis,
  Zettlemoyer, and Stoyanov}]{DBLP:journals/corr/abs-1907-11692}
Yinhan Liu, Myle Ott, Naman Goyal, Jingfei Du, Mandar Joshi, Danqi Chen, Omer
  Levy, Mike Lewis, Luke Zettlemoyer, and Veselin Stoyanov. 2019.
\newblock \href {http://arxiv.org/abs/1907.11692} {Roberta: {A} robustly
  optimized {BERT} pretraining approach}.
\newblock \emph{CoRR}, abs/1907.11692.

\bibitem[{Liu et~al.(2020)Liu, Xiong, Sun, and Liu}]{liu-etal-2020-fine}
Zhenghao Liu, Chenyan Xiong, Maosong Sun, and Zhiyuan Liu. 2020.
\newblock \href {https://doi.org/10.18653/v1/2020.acl-main.655} {Fine-grained
  fact verification with kernel graph attention network}.
\newblock In \emph{Proceedings of the Annual Meeting of the Association for
  Computational Linguistics}, pages 7342--7351.

\bibitem[{Louizos et~al.(2018)Louizos, Welling, and Kingma}]{hard-concrete}
Christos Louizos, Max Welling, and Diederik~P. Kingma. 2018.
\newblock \href {https://openreview.net/forum?id=H1Y8hhg0b} {Learning sparse
  neural networks through ${L}_0$ regularization}.
\newblock In \emph{International Conference on Learning Representations}.

\bibitem[{Lundberg and Lee(2017)}]{shap_nips}
Scott~M Lundberg and Su-In Lee. 2017.
\newblock \href
  {https://proceedings.neurips.cc/paper/2017/file/8a20a8621978632d76c43dfd28b67767-Paper.pdf}
  {A unified approach to interpreting model predictions}.
\newblock In \emph{Advances in Neural Information Processing Systems},
  volume~30.

\bibitem[{Lyu et~al.(2022)Lyu, Apidianaki, and Callison-Burch}]{survey2022}
Qing Lyu, Marianna Apidianaki, and Chris Callison-Burch. 2022.
\newblock \href {http://arxiv.org/abs/2209.11326} {Towards faithful model
  explanation in nlp: A survey}.
\newblock \emph{arXiv preprint}, arXiv:2209.11326.

\bibitem[{Meister et~al.(2021)Meister, Lazov, Augenstein, and
  Cotterell}]{DBLP:conf/acl/MeisterLAC20}
Clara Meister, Stefan Lazov, Isabelle Augenstein, and Ryan Cotterell. 2021.
\newblock \href {https://doi.org/10.18653/v1/2021.acl-short.17} {Is sparse
  attention more interpretable?}
\newblock In \emph{Proceedings of the 59th Annual Meeting of the Association
  for Computational Linguistics and the 11th International Joint Conference on
  Natural Language Processing}, pages 122--129.

\bibitem[{Mudrakarta et~al.(2018)Mudrakarta, Taly, Sundararajan, and
  Dhamdhere}]{layer-integrated-gradient}
Pramod~Kaushik Mudrakarta, Ankur Taly, Mukund Sundararajan, and Kedar
  Dhamdhere. 2018.
\newblock \href {https://doi.org/10.18653/v1/P18-1176} {Did the model
  understand the question?}
\newblock In \emph{Proceedings of the Annual Meeting of the Association for
  Computational Linguistics}, pages 1896--1906.

\bibitem[{Ostrowski et~al.(2021)Ostrowski, Arora, Atanasova, and
  Augenstein}]{politihop}
Wojciech Ostrowski, Arnav Arora, Pepa Atanasova, and Isabelle Augenstein. 2021.
\newblock \href {https://doi.org/10.24963/ijcai.2021/536} {Multi-hop fact
  checking of political claims}.
\newblock In \emph{Proceedings of the International Joint Conference on
  Artificial Intelligence}, pages 3892--3898.

\bibitem[{Paranjape et~al.(2020)Paranjape, Joshi, Thickstun, Hajishirzi, and
  Zettlemoyer}]{informationbottleneck}
Bhargavi Paranjape, Mandar Joshi, John Thickstun, Hannaneh Hajishirzi, and Luke
  Zettlemoyer. 2020.
\newblock \href {https://doi.org/10.18653/v1/2020.emnlp-main.153} {An
  information bottleneck approach for controlling conciseness in rationale
  extraction}.
\newblock In \emph{Proceedings of the Conference on Empirical Methods in
  Natural Language Processing}, pages 1938--1952.

\bibitem[{Popat et~al.(2018)Popat, Mukherjee, Yates, and
  Weikum}]{popat-etal-2018-declare}
Kashyap Popat, Subhabrata Mukherjee, Andrew Yates, and Gerhard Weikum. 2018.
\newblock \href {https://doi.org/10.18653/v1/D18-1003} {{D}e{C}lar{E}:
  Debunking fake news and false claims using evidence-aware deep learning}.
\newblock In \emph{Proceedings of the Conference on Empirical Methods in
  Natural Language Processing}, pages 22--32.

\bibitem[{Ribeiro et~al.(2016)Ribeiro, Singh, and Guestrin}]{lime}
Marco~Tulio Ribeiro, Sameer Singh, and Carlos Guestrin. 2016.
\newblock \href {https://doi.org/10.1145/2939672.2939778} {"why should i trust
  you?": Explaining the predictions of any classifier}.
\newblock In \emph{Proceedings of the ACM SIGKDD International Conference on
  Knowledge Discovery and Data Mining}, page 1135–1144.

\bibitem[{Sabour et~al.(2017)Sabour, Frosst, and Hinton}]{capsule}
Sara Sabour, Nicholas Frosst, and Geoffrey~E. Hinton. 2017.
\newblock \href
  {https://proceedings.neurips.cc/paper/2017/file/2cad8fa47bbef282badbb8de5374b894-Paper.pdf}
  {Dynamic routing between capsules}.
\newblock In \emph{Proceedings of the International Conference on Neural
  Information Processing Systems}, page 3859–3869.

\bibitem[{Shu et~al.(2019)Shu, Cui, Wang, Lee, and Liu}]{defend-kdd}
Kai Shu, Limeng Cui, Suhang Wang, Dongwon Lee, and Huan Liu. 2019.
\newblock \href {https://doi.org/10.1145/3292500.3330935} {Defend: Explainable
  fake news detection}.
\newblock In \emph{Proceedings of the ACM SIGKDD International Conference on
  Knowledge Discovery \& Data Mining}, page 395–405.

\bibitem[{Si et~al.(2021)Si, Zhou, Li, Shi, and He}]{si2021}
Jiasheng Si, Deyu Zhou, Tongzhe Li, Xingyu Shi, and Yulan He. 2021.
\newblock \href {https://doi.org/10.18653/v1/2021.acl-long.128} {Topic-aware
  evidence reasoning and stance-aware aggregation for fact verification}.
\newblock In \emph{Proceedings of the Annual Meeting of the Association for
  Computational Linguistics and the International Joint Conference on Natural
  Language Processing}, pages 1612--1622.

\bibitem[{Si et~al.(2022)Si, Zhu, and Zhou}]{si2022exploring}
Jiasheng Si, Yingjie Zhu, and Deyu Zhou. 2022.
\newblock \href {https://arxiv.org/abs/2212.01060} {Exploring faithful
  rationale for multi-hop fact verification via salience-aware graph learning}.
\newblock \emph{arXiv preprint arXiv:2212.01060}.

\bibitem[{Sundararajan et~al.(2017)Sundararajan, Taly, and Yan}]{intgrad}
Mukund Sundararajan, Ankur Taly, and Qiqi Yan. 2017.
\newblock \href {https://proceedings.mlr.press/v70/sundararajan17a.html}
  {Axiomatic attribution for deep networks}.
\newblock In \emph{Proceedings of the International Conference on Machine
  Learning}, volume~70, pages 3319--3328.

\bibitem[{Wiegreffe and Pinter(2019)}]{DBLP:conf/emnlp/WiegreffeP19}
Sarah Wiegreffe and Yuval Pinter. 2019.
\newblock \href {https://doi.org/10.18653/v1/D19-1002} {Attention is not not
  explanation}.
\newblock In \emph{Proceedings of the 2019 Conference on Empirical Methods in
  Natural Language Processing and the 9th International Joint Conference on
  Natural Language Processing}, pages 11--20.

\bibitem[{Wu et~al.(2021)Wu, Rao, Lan, Sun, and Qi}]{wu-etal-2021-unified}
Lianwei Wu, Yuan Rao, Yuqian Lan, Ling Sun, and Zhaoyin Qi. 2021.
\newblock \href {https://doi.org/10.18653/v1/2021.acl-long.5} {Unified
  dual-view cognitive model for interpretable claim verification}.
\newblock In \emph{Proceedings of the Annual Meeting of the Association for
  Computational Linguistics and the International Joint Conference on Natural
  Language Processing}, pages 59--68.

\bibitem[{Wu et~al.(2020)Wu, Rao, Zhao, Liang, and Nazir}]{wu-etal-2020-dtca}
Lianwei Wu, Yuan Rao, Yongqiang Zhao, Hao Liang, and Ambreen Nazir. 2020.
\newblock \href {https://doi.org/10.18653/v1/2020.acl-main.97} {{DTCA}:
  Decision tree-based co-attention networks for explainable claim
  verification}.
\newblock In \emph{Proceedings of the Annual Meeting of the Association for
  Computational Linguistics}, pages 1024--1035.

\bibitem[{Yan et~al.(2022)Yan, Gui, and He}]{yan2022hierarchical}
Hanqi Yan, Lin Gui, and Yulan He. 2022.
\newblock \href {https://arxiv.org/abs/2202.09792} {Hierarchical interpretation
  of neural text classification}.
\newblock \emph{arXiv preprint arXiv:2202.09792}.

\bibitem[{Yang et~al.(2019)Yang, Pentyala, Mohseni, Du, Yuan, Linder, Ragan,
  Ji, and Hu}]{xfact-www19}
Fan Yang, Shiva~K. Pentyala, Sina Mohseni, Mengnan Du, Hao Yuan, Rhema Linder,
  Eric~D. Ragan, Shuiwang Ji, and Xia~(Ben) Hu. 2019.
\newblock \href {https://doi.org/10.1145/3308558.3314119} {Xfake: Explainable
  fake news detector with visualizations}.
\newblock In \emph{The World Wide Web Conference}, page 3600–3604.

\bibitem[{Yu et~al.(2021)Yu, Zhang, Chang, and Jaakkola}]{yu2021understanding}
Mo~Yu, Yang Zhang, Shiyu Chang, and Tommi Jaakkola. 2021.
\newblock \href {https://arxiv.org/abs/2110.13880v1} {Understanding
  interlocking dynamics of cooperative rationalization}.
\newblock In \emph{Advances in Neural Information Processing Systems},
  volume~34.

\bibitem[{Zhang et~al.(2021)Zhang, Rudra, and Anand}]{predictagain2021}
Zijian Zhang, Koustav Rudra, and Avishek Anand. 2021.
\newblock \href {https://doi.org/10.1145/3437963.3441758} {Explain and predict,
  and then predict again}.
\newblock In \emph{Proceedings of the ACM International Conference on Web
  Search and Data Mining}, page 418–426.

\bibitem[{Zhao et~al.(2020)Zhao, Xiong, Rosset, Song, Bennett, and
  Tiwary}]{transformer-xh}
Chen Zhao, Chenyan Xiong, Corby Rosset, Xia Song, Paul Bennett, and Saurabh
  Tiwary. 2020.
\newblock \href {https://openreview.net/forum?id=r1eIiCNYwS} {Transformer-xh:
  Multi-evidence reasoning with extra hop attention}.
\newblock In \emph{International Conference on Learning Representations}.

\bibitem[{Zhou et~al.(2020)Zhou, Hu, Zhang, Liang, Sun, Xiong, and
  Tang}]{NEURIPS2020_zhou}
Wangchunshu Zhou, Jinyi Hu, Hanlin Zhang, Xiaodan Liang, Maosong Sun, Chenyan
  Xiong, and Jian Tang. 2020.
\newblock \href
  {https://proceedings.neurips.cc/paper/2020/file/4be2c8f27b8a420492f2d44463933eb6-Paper.pdf}
  {Towards interpretable natural language understanding with explanations as
  latent variables}.
\newblock In \emph{Advances in Neural Information Processing Systems},
  volume~33, pages 6803--6814.

\end{thebibliography}
\bibliographystyle{acl_natbib}

\end{document}